# A constraint satisfaction framework for decision under uncertainty

**Hélène Fargier, Jérôme Lang**
IRIT - Université Paul Sabatier
31062 Toulouse Cedex (France)
{fargier, lang}@irit.fr

**Roger Martin-Clouaire, Thomas Schiex**
INRA, BP 27
31326 Castanet Cedex (France)
{rmc,tschiex}@toulouse.inra.fr

## Abstract

*The Constraint Satisfaction Problem (CSP) framework offers a simple and sound basis for representing and solving simple decision problems, without uncertainty. This paper is devoted to an extension of the CSP framework enabling us to deal with some decisions problems under uncertainty. This extension relies on a differentiation between the agent-controllable decision variables and the uncontrollable parameters whose values depend on the occurrence of uncertain events. The uncertainty on the values of the parameters is assumed to be given under the form of a probability distribution. Two algorithms are given, for computing respectively decisions solving the problem with a maximal probability, and conditional decisions mapping the largest possible amount of possible cases to actual decisions.*

## 1 Introduction

Decision making is primarily a matter of choosing between alternatives that most commonly are expressed implicitly. Thus solving a decision problem amounts to generate the option(s) that is(are) most appropriate with respect to the specification of the decision problem at hand. Different decision problems can be characterized along two discriminating features:

- decisions are made at a single time point (although they may have a temporal structure *i.e.*, form a plan) or are elaborated in sequence of steps, each being enriched by information resulting from the previous ones;
- the requirements (constraints, criteria) that implicitly define the alternatives are uncertainty-free or not. Requirements are uncertain if they vary depending on the circumstances, the occurrence of which is incompletely known (uncertain).

Sequential decisions and uncertain requirements increase the complexity of the decision problem, the most complex case being the joint situation. This paper is devoted to the case of single-instant decisions and uncertain requirements and proposes an approach of the problem in the framework of constraint satisfaction.

So far the only decision problem tackled within CSP (Constraint Satisfaction Problem) approaches concern the simplest case: single-instant decision problem with no uncertainty at all. In order to cast the addressed decision problem in a CSP framework it is essential to distinguish between two types of unknowns that are called *parameters* and *decision variables* respectively. Parameters are *uncontrollable unknowns*, *i.e.*, the value taken by a parameter is a matter of occurrence of an event that cannot be controlled (not even influenced) by the decision maker (also referred to as the agent). The value of a parameter is imposed by the external world and the agent may have only partial knowledge about what this value might be. By contrast, the assignment of the decision variables is what the agent wants to decide upon. Failing to differentiate between parameters and decision variables may yield nonsensical results in a classical CSP approach because a particular value of a parameter can restrict the range of allowed (satisfactory) values for the decision variables whereas the converse is physically impossible. The key issue is therefore to develop a CSP framework and resolution algorithms that provide for the uncontrollable/controllable dichotomy in the set of unknowns.

This paper addresses an extension of the CSP framework, namely *probabilistic CSP*, involving both parameters and decision variables, the uncertainty on the values of the parameters being represented by a probability distribution. A formal representation framework is defined and two algorithms relying on an assumption of independence of the parameters are described. In this paper we consider successively two assumptions concerning the agent's awareness of the paremeter values (the state of the world) at the time the decision must imperatively be made (deadline for acting):

- **(NK)** ("no more knowledge"): the agent will never learn anything new about the actual state of the world before the deadline; all it will ever know is already encoded by the probability distribution. For this case we propose an algorithm which gives an actual, unconditional decision, that is most likely to be suitable.
- **(CK)** ("complete knowledge"): the actual world will be completely revealed before the deadline is reached (possibly, *just* before), so that it it useful to the agent to compute "off-line" a ready-to-use conditional decision, that the agent will be able to instantiate "on-line",

168   Fargier, Lang, Martin-Clouaire, and Schiex

as soon as it knows what the actual world is. For this purpose we have developed an "anytime" resolution algorithm that provides a set of decisions with their conditions of applicability, together with the likelihood of occurrence of these conditions.

The intermediate case where some partial knowledge about the state of the world may be learned is not considered in this paper. See [5] for further discussion on this point.

In Section 2 we define probabilistic CSPs and several types of solutions (corresponding to decisions); then we propose algorithms dealing with the problem of computing optimal decisions under the assumptions (**NK**) (Section 3) and (**CK**) (Section 4).

## 2 Probabilistic constraint satisfaction problems

### 2.1 Preliminary definitions and notations

A classical *constraint satisfaction problem* is a triple $P = \langle X, D, C \rangle$, where $X$ is a set of *variables*, each of which has its possible values in a domain (supposed here finite) $D_i$ (with $D = \times_i D_i$), and $C$ is a set of *constraints*. Each constraint $C_i \in C$ involves a set of variables noted $V(C_i)$ and is defined by a subset of the Cartesian product of the domains of the variables in $V(C_i)$. This subset, noted $C_i$ as the constraint itself, gives the set of all the possible assignments of the variables in $V(C_i)$: the constraint $C_i$ is satisfied by an assignment of the variables in $V(C_i)$ iff this assignment belongs to the set of admissible tuples of $C_i$. A solution of the CSP is an assignment of values to all the variables such that all the constraints are satisfied. The set of all the solutions of a CSP $P$ is noted $Sol(P)$.

In the rest of the paper, assignments of values to a set of variables $Y$ are also considered as tuples of values of the variables of $Y$, *i.e.*, elements of the Cartesian product of the domains of the variables in $Y$. The concatenation of two assignments $a_1$ and $a_2$ of variables in $V_1$ and $V_2$ such that $V_1 \cap V_2 = \emptyset$ is noted $(a_1, a_2)$; it is an assignment of $V_1 \cup V_2$. The projection of an assignment $a$ of a set of variables $Y$ on a set of variables $Z \subseteq Y$, noted $t_{\downarrow Z}$ is simply the tuples of the values of the variables of $Z$ in $t$. This notion (and notation) is extended to a set of tuples: the projection $R_{\downarrow Z}$ on $Z$ of a subset $R$ of the Cartesian product of the domains of the variables of $Y$ is the set of the projections on $Z$ of all the assignments in $R$.

### 2.2 Probabilistic CSPs: definitions

Roughly speaking, a probabilistic CSP is a CSP equipped with a partition between (controllable) decision variables and (uncontrollable) parameters, and a probability distribution over the possible values of the parameters.

**Definition 1** *A* probabilistic CSP *is a 6-uple* $\mathcal{P} = \langle \Lambda, W, X, D, C, pr \rangle$ *where:*

- $\Lambda = \{\lambda_1, \ldots, \lambda_p\}$ *is a set of parameters;*
- $W = W_1 \times \cdots \times W_p$, *where $W_i$ is the domain of $\lambda_i$;*
- $X = \{x_1, \ldots, x_n\}$ *is a set of decision variables;*
- $D = D_1 \times \cdots \times D_n$, *where $D_i$ is the domain of $x_i$;*
- $C$ *is a set of constraints, each of them involving at least one decision variable.*
- $pr : W \to [0, 1]$ *is a probability distribution over the parameter assignments.*

*Constraints are defined in the same way as in classical CSP. We note $X(C_i)$ (resp. $\Lambda(C_i)$) the set of variables (resp. parameters) involved in a constraint $C_i$.*

A complete assignment of the parameters (resp. of the decision variables) will be called a *world* (resp. a *decision*) and will be generally denoted by $\omega$ (resp. by $d$). A partial world (resp. a partial decision) will be an assignment of a subset of the parameters (resp. of the decision variables).

The subset of $C$ containing the constraints involving no parameters will be denoted by $C^*$; the other constraints in $C$ (involving at least a parameter) restrict the allowed values of some decision variables, *dependently of the values of some parameters*: they will be called *parameterized* constraints. Obviously, if $\Lambda = \emptyset$, $\mathcal{P}$ is a classical CSP.

We assume that the constraints of $C$ involve at least a decision variable since the available information about the actual values of parameters is completely encoded by $pr$ — if a tuple of parameter assignments is impossible then the probability of any world extending this tuple is 0.

We have not yet discussed how the probability distribution on worlds is specified. Clearly, it is not reasonable to assume that the input contains the explicit specification of $pr(\omega)$ for each $\omega$. Hence, generally $pr$ will be given in a much more concise way. The simplest case occurs when parameters are mutually independent ($pr$ is then specified only by an individual probability distributions $pr_{\lambda_i}$ for each parameter $\lambda_i$). A more complex case consists in structuring the parameters in a Bayesian network: the computation of $pr$ requires then the propagation of probabilities through the network. Thus, $pr$ will generally be given *implicitly* rather than explicitly, and will generally require some computation in order to be available. From now on we choose to ignore this step: the computation of $pr$ is taken for granted.

**Definition 2 (possible worlds)** *A world $\omega$ of $W$ such that $pr(\omega) > 0$ is a* possible world. *The set of all possible worlds is denoted by $Poss(\mathcal{P})$.*

Let $a$ be an assignment of a subset of the variables (where *variables* is here the general terminology for both parameters and decision variables). We define the *reduction* of a constraint $C$ by assignment $a$ as the set of assignments of the unassigned variables compatible with $a$ according to $C$:

**Definition 3** *Let $C$ be a constraint of $C$ involving a set $V(C) = \Lambda(C) \cup X(C)$ of variables and parameters. The reduction of $C$ by an assignment $a$ of $V' \subseteq V(C)$ is the constraint $Reduce(C, a)$ on $V(C) - V'$ defined by the tuples $\{b \in D_{\downarrow V(C) - V'} \mid (b, a)_{\downarrow V(C)} \in C\}$.*

*We note $C[a] = \{Reduce(C, a), C \in C\}$ the reduction of $C$ by $a$.*

This notion of reduction is especially meaningful when $a$ is either a world $\omega$ or a decision $d$. The reduction of $C$ by a world $\omega$ is a set of classical decision constraints (involving decision variables only); it defines the decisions which are suitable if the world is $\omega$. Namely, to each world $\omega$ we associate the uncertainty-free decision problem $\langle X, D, C[\omega]\rangle$. Each of these classical CSPs will be called a *candidate problem*. One of them (and only one) is the *actual problem* $\langle X, D, C[\omega^*]\rangle$, corresponding to the *actual world* $\omega^*$ (but since there is generally more than one possible world, the agent does not know which one is the actual one).

**Definition 4** *The set of **candidate problems** induced by $\mathcal{P}$ is defined by $Candidates(\mathcal{P}) = \{\langle X, D, C[\omega]\rangle \mid \omega \in W)\}$.*

Note that among the CSPs of $Candidates(\mathcal{P})$, some may be inconsistent, which means that the actual problem may be inconsistent.

Dually, the reduction of $C$ by a decision $d$ yields a CSP $\langle \Lambda, W, C[d]\rangle$ involving parameters only, whose solutions are the worlds for which $d$ is a suitable decision. These worlds are said to be *covered* by $d$. Note that, obviously, $\omega$ is a solution of $\langle \Lambda, W, C[d]\rangle$ iff $d$ is a solution of $\langle X, D, C[\omega]\rangle$ iff $(\omega, d)$ is a solution of $\langle \Lambda \cup X, W \times D, C\rangle$.

Among the possible worlds, those which are covered by at least a decision are called *good worlds* — and the others are called *bad worlds*. Equivalently, $\omega$ is good iff $\langle X, D, C[\omega]\rangle$ is consistent.

**Definition 5**
$Good(\mathcal{P}) = \{\omega \in Poss(\mathcal{P}) \mid \exists d \in D, (\omega, d) \text{ satisfies } C\}$
$Bad(\mathcal{P}) = Poss(\mathcal{P}) \setminus Good(\mathcal{P})$

Now, for each possible decision $d$ we can compute the probability that it is a suitable decision, *i.e.* that it covers the actual world.

**Definition 6 (probability that a decision is a solution)**
*The probability that a given decision $d$ is a solution of the actual problem is the probability of the set of worlds it covers, i.e. $PS(d) = Pr(Sol(\langle \Lambda, W, C[d]\rangle))$.*

Notice that $PS$ *is not* a probability distribution[1].

Now, we can compute the probability that the actual world can be covered: this is the probability of consistency of the actual problem.

**Definition 7 (probability of consistency)** *To a probabilistic CSP $\mathcal{P}$ we associate the probability that the actual problem is consistent, i.e.*

$$P_{Cons}(\mathcal{P}) = Pr(Good(\mathcal{P}))$$

*If $P_{Cons}(\mathcal{P}) = 1$ then $\mathcal{P}$ will be said consistent.*

---
[1]It is actually a contour function in the sense of Dempster-Shafer theory. We omit details due to lack of space (see [5]).

*Example: (strongly modified from [14]). Consider a dinner to be organized, to which each of the three guests (Grandgousier, Gargantua and Pantagruel) is not sure to come. The problem is to choose a wine and a meal according to four constraints, among which three depends on the presence of each guest. The probabilistic CSP $\mathcal{P}$ corresponding to this problem is defined by:*

- *Decision variables:* $X = \{x_1, x_2\}$; $D_1 = \{White, Red\}$; $D_2 = \{Turkey, Beef, Fish\}$
- *Parameters:* $\Lambda = \{\lambda_1, \lambda_2, \lambda_3\}$; $W_1 = W_2 = W_3 = \{comes, \neg comes\}$ *(these two values are abbreviated c and $\neg c$ in the rest of the paper).*
- *Constraints:* $C = \{C_1, C_2, C_3, C_4\}$;

Cuisine rules

$C_1:$

| $x_1$ | $x_2$ |
|---|---|
| W | F |
| R | B |
| W | T |
| R | T |

Grandgousier (beef)

$C_2:$

| $\lambda_1$ | $x_2$ |
|---|---|
| c | T |
| c | F |
| $\neg c$ | T |
| $\neg c$ | F |
| $\neg c$ | B |

Gargantua (white wine)

$C_3:$

| $\lambda_2$ | $x_1$ |
|---|---|
| c | R |
| $\neg c$ | R |
| $\neg c$ | W |

Pantagruel (vegetarian)

$C_4:$

| $\lambda_3$ | $x_2$ |
|---|---|
| c | F |
| $\neg c$ | T |
| $\neg c$ | F |
| $\neg c$ | B |

*The participation of the different guests are mutually independent. The probability that Grandgousier comes is $0.6$ and respectively $0.9$ and $0.5$ for Gargantua and Pantagruel.*

*In this example, all worlds are possible ($Poss(\mathcal{P}) = W$). If we consider world $\omega_1 = (c, \neg c, c)$ whose probability is $0.03$, the reduction of the constraints are: $Reduce(C_1, \omega_1) = C_1$, $Reduce(C_2, \omega_1) = \{T, F\}$, $Reduce(C_3, \omega_1) = \{R, W\}$ and $Reduce(C_3, \omega_1) = \{F\}$. The CSP $\langle X, D, C[\omega_1]\rangle$ that corresponds to $\omega_1$ has a single solution $(W, F)$ and thus $\omega_1$ is a good world. The decision $(W, F)$ covers $3$ more worlds $((\neg c, \neg c, c), (c, \neg c, \neg c)$ and $(\neg c, \neg c, \neg c))$ which finally give a probability $PS((W, F)) = 0.1$ (since the worlds covered by $(W, f)$ are exactly those where Gargantua does not come).*

*In the world $\omega_2 = (c, c, c)$, whose probability is $0.27$, the reduction of the constraints are: $Reduce(C_1, \omega_2) = C_1$, $Reduce(C_2, \omega_2) = \{T, F\}$, $Reduce(C_3, \omega_2) = \{R\}$ and $Reduce(C_3, \omega_2) = \{F\}$ which defines an inconsistent CSP. Thus, $\omega_2$ is a bad world. The only other bad world is $\omega_3 = (\neg c, c, c)$ with probability $0.18$. Therefore, $Bad(\mathcal{P}) = \{(c, c, c), (\neg c, c, c)\}$, $Pr(Bad(\mathcal{P})) = 0.27 + 0.18 = 0.45$ and $P_{Cons}(\mathcal{P}) = 0.55$.*

There is a number of interesting particular cases of probabilistic CSPs obtained by some simplifying assumptions. A first assumption, already evoked in Section 2.1, is the mutual independence of parameters: in this case $pr$ is specified only by individual probability distributions[2].

Another particular case is illustrated by our previous example: we have a problem where the relevance of some constraints is uncertain, and every uncertain constraint $c_i$ is encoded by a parameterized constraint $C_i$ linking the decision variables of $c_i$ to one parameter $\lambda_i$, whose domain has two values (in our example, *comes* and $\neg comes$), corresponding respectively to the relevance and the irrelevance of $c_i$ to the actual problem. This simplifying assumption is denoted by $(U)$. If $\{c_1, ..., c_q\}$ denotes the set of these uncertain constraints, then it can be shown that $Candidates(\mathcal{P})$ is a relaxation lattice in the sense of Freuder [6], equipped with the probability distribution obviously induced by the distribution on worlds: namely, the set of candidates problems are obtained by taking the union of $\mathcal{C}^*$ and of any subset of $\{c_i, i = 1 \ldots q\}$. Moreover when the $c_i$'s are independently relevant, $\mathcal{P}$ is strongly consistent (see Definition 13) iff the top of the lattice, namely $\{c_i, i = 1 \ldots q\} \cup \mathcal{C}^*$ is consistent in the classical sense. A more general class of probabilistic CSPs occurs when each parameterized constraint involves exactly one parameter. Each parameterized constraint corresponds thus to an disjunctive family of classical decision constraints (one for each possible value of the involved parameter). We denote by $(F)$ this simplifying assumption.

---
[2]Hence, the problem can be stated in such a way that $Poss(\mathcal{P}) = W$, simply by removing impossible values from the domain.

## 2.3 Conditional decisions as solutions

A conditional decision will ideally associate to each possible world a decision satisfying the corresponding decision problem. Since some possible worlds may induce an inconsistent CSP, a weaker request consists in giving a partial conditional decision, *i.e.*, defined on a subset of $W$ (ideally on $Good(\mathcal{P})$).

**Definition 8 (conditional decisions)** *A complete conditional decision $s$ is a map from $Poss(\mathcal{P})$ to $D$. A partial conditional decision $s$ is a map from a subset $W(s)$ of $Poss(\mathcal{P})$ to $D$. A (partial or complete) conditional decision is sound iff $\forall \omega \in W(s), s(\omega)$ covers $\omega$.*

In the rest of the paper, we refer to "conditional decisions" as to partial or complete conditional decisions. Clearly, complete conditional decisions generalize decisions as defined in Section 2.2 (the latter are called *pure decisions* as opposed to conditional decisions). Namely, a conditional decision which is constant over $Poss(\mathcal{P})$ corresponds to a (pure) decision. Conditional decisions will play the role of solutions for probabilistic CSPs (ideally, a solution is a complete sound conditional decision). As for (pure) decisions, a conditional decision has a probability to cover the actual world:

**Definition 9** *The probability that a conditional decision $s$ will yield a solution to the actual problem is defined by:*

$$PS(s) = \sum_{\substack{\omega \in W(s) \\ s(\omega) \text{ covers } \omega}} pr(\omega)$$

The following simple results are easy to prove:

**Proposition 1**
*(i) for any conditional decision $s$, $PS(s) \leq P_{Cons}(\mathcal{P})$.*
*(ii) there exists a $s$ such that $PS(s) = P_{Cons}(\mathcal{P})$.*
*(iii) $\mathcal{P}$ is consistent iff there exists a $s$ such that $PS(s) = 1$.*

**Definition 10 (optimal conditional decisions)** *A conditional decision (partial or complete) $s$ is optimal for $\mathcal{P}$ iff $PS(s)$ is maximum (or equivalently iff $PS(s) = P_{Cons}(\mathcal{P})$).*

Thus, one may compute an optimal conditional decision *off line* and use it later on, in real time, when the actual world will be known. This assumes that this actual world will be known before the decision has to be taken ((**CK**)). Clearly, under the other extreme assumption ((**NK**)), it is useless to look for a conditional decision: consequently, in this case one has to look for a *pure decision* rather than for a conditional one. A reasonable strategy is then to maximize the probability that this pure decision will work:

**Definition 11 (optimal pure decisions)** *A pure decision $d$ is optimal iff $PS(d)$ is maximum. We let $P_{SPD}(\mathcal{P}) = PS(d)$ where $d$ is an optimal pure decision.*

$P_{SPD}(\mathcal{P})$ is the maximum probability of success of a Pure Decision. Obviously, $P_{SPD}(\mathcal{P}) \leq P_{Cons}(\mathcal{P})$. The ideal case is when there is a pure decision covering all good worlds:

**Definition 12 (universal decisions)** *A pure decision $d$ is a universal decision iff $PS(d) = P_{Cons}(\mathcal{P})$.*

Thus, a universal decision will work for any possible world whose associated candidate problem is consistent and thus it is an optimal decision. Clearly, when there exists a universal decision, it is useless to look for a conditional decision.

**Definition 13 (strong universal decisions)** *A pure decision $d$ is a strong universal decision of $\mathcal{P}$ iff $PS(d) = 1$. $\mathcal{P}$ is strongly consistent iff it has a strong universal decision.*

A strong universal decision is a universal decision covering *all* possible worlds. The case where $\mathcal{P}$ is strongly consistent is the ideal one. Obviously, strong consistency implies consistency.

*Example*: *using the same probabilistic CSP, we may consider the following conditional decision $s$, which is optimal since $PS(s) = P_{Cons}(\mathcal{P}) = 0.55$.*

| $\lambda_1$ | $\lambda_2$ | $\lambda_3$ | Decision |
|---|---|---|---|
| $c$ | $c$ | $c$ | $Bad$ |
| $c$ | $c$ | $\neg c$ | $(R, T)$ |
| $c$ | $\neg c$ | $\neg c$ | $(R, T)$ |
| $c$ | $\neg c$ | $c$ | $(W, F)$ |
| $\neg c$ | $c$ | $c$ | $Bad$ |
| $\neg c$ | $c$ | $\neg c$ | $(R, T)$ |
| $\neg c$ | $\neg c$ | $\neg c$ | $(R, T)$ |
| $\neg c$ | $\neg c$ | $c$ | $(W, F)$ |

*The decision $(R, T)$ is the only optimal pure decision ($PS((R, T)) = 0.5$) but is not universal. However, if the probability that Pantagruel ($\lambda_3$) comes were 0, $(R, T)$ would be a strong universal decision.*

## 3 Searching for an optimal pure decision

In both Sections 3 and 4 we assume mutual independence of parameters, for the sake of simplicity; however this as-sumption could be relaxed provided that the computation of the probability of a set of worlds is taken for granted.

Furthermore, in this Section (only) we make the assumption (**NK**) that the agent will never have any further knowledge about the actual world before acting (all it knows is already encoded in $pr$) and consequently, all it can execute is a pure decision. Now, the best the agent can require from the solver is an *optimal* pure decision. As it may be computationally costly to compute an optimal one, especially if the agent has a deadline, we propose an "anytime" algorithm which computes an approximately optimal pure decision if stopped before its natural stop (the longer the algorithm runs, the better the pure decision), and eventually gives an optimal one if it runs until its natural stop.

Our approach to the problem of maximizing $PS(d)$ consists in using a Depth First Branch and Bound algorithm. For the sake of simplicity, consider that variables are instantiated in a prescribed order, say $(x_1, \ldots, x_n)$ and that all the constraint are binary[3]. The root of the tree is the empty instantiation. Intermediate nodes denote partial decisions $(d_1, \ldots, d_i)$. Leaves represent instantiations of $X$, *i.e.*, pure decisions $d = (d_1, \ldots, d_n)$. In a depth first exploration of the tree, we keep track of the leaves $d$ maximizing $PS(d)$.

Each time a new variable decision $d_i$ is assigned to variable $x_i$, all the constraints $C \in \mathcal{C}$ connecting this variable $x_i$ to an unassigned decision variable $x_j$ or a parameter $\lambda_k$ will be used as in the classical *Forward-checking* algorithm [8] to remove all the values incompatible with the assignment $x_i = d_i$ from the domains of $x_j$ (resp. $\lambda_k$): each domain $D_j$ (resp. $W_k$) is simply replaced by $(D_j \cap Reduce(C, (d_i)))$ (resp $(W_k \cap Reduce(C, (d_i)))$ if $C$ involves a parameter). Updating the domain of a variable $v$ using constraint $C$ after the assignment of decision $d_i$ to $x_i$ is performed by the procedure FORWARDCHECK($v, C, d_i$) in Algorithm 1, line 2.

After forward-checking, the Cartesian product of the remaining values in the domains of all the parameters define the worlds which are compatible with the current partial decision: all the worlds that are obviously incompatible have been removed. The cumulated probabilities of the remaining worlds give an upper bound $\widehat{PS}(d)$ on the probability of the best complete decision among the descendants of the current node since only incompatible worlds have been removed. This bound is easily computed if independence is assumed: the bound is simply the product on all domains of the sums of the probabilities of the remaining values of each domain. When a complete decision $d$ is reached, since all the constraint have been propagated, the worlds remaining *are* the worlds covered by the decision: $\widehat{PS}(d) = PS(d)$.

When does backtrack occur ? First, if the domain of a decision variable or a parameter becomes empty, then we know that no world is covered by our current partial decision and we may therefore backtrack, reconsider the value of the last decision variable assigned, restore the domains modified due to the last assignment to their previous values and go on. Further cutoffs can be obtained using our upper bound and a very simple lower bound on the best probability. This lower bound is simply the probability $PS(d)$ associated to the best decision found so far and is embodied in the algorithm in a threshold $\alpha$, initialized to 0 and updated each time a decision $d$ such that $PS(d) > \alpha$ is reached. Whenever the two bounds meet, we know that the best probability of any complete decision that can be reached from this node is worse than the best probability found up to now and backtrack occurs, as previously.

---

[3]This implies that we are in case (F) since a constraint involves at least one decision variable. The algorithm is easily extended to handle larger arities as long as assumption (F) holds.

The search will stop when no more node can be created. It is successful if a leaf has been reached, and the best decision among those which have been reached is optimal.

**Function** SEARCH($d_1, \ldots, d_i, p_i$)
**if** $i = n$ **then** $\alpha := p_i$           {Update upper bound}
**else**
    Choose a variable $x_{i+1}$
    **for** $d_{i+1} \in D_{i+1}$ **do**
        SAVEDOMAINS($\Lambda \cup X$)
        **for** $C \in \mathcal{C}$ s.t. $x_{i+1} \in V(C)$
        and $V(C) \not\subseteq \{x_1, \ldots, x_{i+1}\}$ **do**
1             Let $\{v_j\} = V(C) - \{x_{i+1}\}$
2             FORWARDCHECK($v_j, C, d_{i+1}$)

3         $p_{i+1} := \prod_{W_k} \sum_{v \in W_k} [pr_{\lambda_k}(v)]$
        **if** $(p_{i+1} > \alpha)$ and $(\forall i, D_i \neq \varnothing)$ **then**
            SEARCH($d_1, \ldots, d_{i+1}, p_{i+1}$)
        RESTOREDOMAINS($\Lambda \cup X$)

Algorithm 1: A Forward-checking-based Depth First Branch and Bound

The algorithm is sketched as Algorithm 1. The function SEARCH should initially be called with an empty partial decision and a probability $p_0$ equal to 1, the initial trivial upper bound on the probability of an optimal decision. The functions SAVEDOMAINS and RESTOREDOMAINS save and restore the domains of the the variables (actually, only modified domains have to be saved). One should note that on line 1 of the algorithm $v_j$ can be a decision variable or a parameter. On line 3, the current upper bound $p_i$ could easily be incrementally updated by only taking into account the parameters whose domains have been modified: the previous probability $p_i$ is simply multiplied by the relative decrease of the probability associated to each modified domain. The algorithm is related to existing extensions of *Forward checking* which have been considered in the context of valued CSP [15] and partial CSP [6].

## 4   Searching for a conditional decision

Clearly, searching for an optimal conditional decision is computationally costly (since in the worst case, its size equals the number of possible worlds); this leads us to look for an algorithm which gives a conditional decision which tends eventually to an optimal one if we let the algorithm run until it stops naturally (the longer we let it run, the higher the probability associated to the current conditional decision). Intuitively, the algorithm incrementally builds a conditional decision which eventually covers a superset of all good worlds. Repeatedly, (i) we pick a new pure decision $d$ (to be added to the current partial conditional decision) that covers at least one possible world among the worlds which are not covered yet, (ii) we compute the set $R$ of worlds that this decision covers and (iii) we subtract this set from the set of worlds which haven't yet been covered (initially, this set has the value $W$, i.e. it contains all worlds). In order to easily compute the worlds covered by $d$ and their probabilities, we make the following two assumptions: first, parameters are mutually independent (again), and second, any constraint of $C$ involves *at most one parameter* (F). Due to assumption (F), the worlds covered by a decision $d$ (i.e., $C[d]$) form a Cartesian product of subsets of the parameter domains. The successive subtractions of these Cartesian products is performed using a technique recently proposed by Freuder and Hubbe [7], called *subdomain subproblem extraction*. Before we describe our algorithm, we first recall this technique of subdomain extraction.

### 4.1 Subdomain extraction

We define an *environment* $E$ as a set of worlds of the form $l_1 \times \cdots \times l_p$, with $\forall k, l_k \subseteq W_k$. An example of environment is $(\lambda_1 \in \{c\}, \lambda_2 \in \{c, \neg c\}, \lambda_3 \in \{\neg c\})$, also written $\{c\} \times \{c, \neg c\} \times \{\neg c\}$ or $\left[\begin{smallmatrix}c\\c\\\neg c\end{smallmatrix}\right]$ when there is no ambiguity on the order of parameters. The set of all possible environments is obviously a lattice (equipped with the inclusion order), whose top is the set of all possible parameter assignments (in the example, $\left[\begin{smallmatrix}c, \neg c\\c, \neg c\\c, \neg c\end{smallmatrix}\right]$) and whose bottom is the empty set. The probability of an environment $E = l_1 \times \cdots \times l_p$, under the independence assumption, is $Pr(E) = \prod_{k=1}^{p} \sum_{v \in l_k} pr(v)$.

Given two environments $E$ and $R$, *sub-domain subproblem extraction* technique [7], decomposes $E$ into a set of *disjoint* sub-environments $Dec(E, R)$ such that all worlds of $E$ belong either to $R$ or to one of the sub-environments of the decomposition. The decomposition is unique if an ordering on the variables is fixed. We give here a modification of Freuder and Hubbe's decomposition algorithm, where we also compute incrementally the probability of each environment generated by the decomposition. This function returns the decomposition of $E$ by $F$ and the probability of the set of worlds of $E$ that are already in $F$ (used in Algorithm 3).

When actually used, the probability $p_E$ of the environment $E$ needed on entry has already been computed since if the procedure DEC is called, either $E = W$ and $p_E = 1$, or $E$ comes from a previous decomposition.

Note that when $E \cap F = \emptyset$ we get $\text{DEC}(E, F) = \{\langle E, p_E\rangle\}$, and when $E \subseteq F$ we get $\text{DEC}(E, F) = \emptyset$.

**Example:** $E = W = \left[\begin{smallmatrix}c, \neg c\\c, \neg c\\c, \neg c\end{smallmatrix}\right]$, $F = \left[\begin{smallmatrix}c\\c\\c\end{smallmatrix}\right]$.

1. $i = 1$; $Rest = \left[\begin{smallmatrix}c\\c, \neg c\\c, \neg c\end{smallmatrix}\right]$; $p_{Rest} = 0.6$, $E' = \left[\begin{smallmatrix}\neg c\\c, \neg c\\c, \neg c\end{smallmatrix}\right]$;

**Function** DEC($\langle E, p_E\rangle, F$)

```
List := ∅; E' := E; p_{E'} = p_E; i := 1
repeat
    Rest := E'
    Rest_{↓{λ_i}} := E'_{↓{λ_i}} - F_{↓{λ_i}}
    p_{Rest} := p_{E'} · p(Rest_{↓{λ_i}})/p(E'_{↓{λ_i}})        {independence}
    if (Rest_{↓{λ_i}} ≠ ∅) then
        Add ⟨Rest, p_{Rest}⟩ to List
        E'_{↓{λ_i}} := E'_{↓{λ_i}} ∩ F_{↓{λ_i}}
        p_{E'} := p_{E'} - p_{Rest}
    i := i + 1
until ((i > p) or (E'_{↓{λ_i}} = ∅))
return ⟨List, p_{E'}⟩
```

Algorithm 2: The decomposition algorithm

   $p_{E'} = 0.4$; List = $\{\langle\left[\begin{smallmatrix}c\\c, \neg c\\c, \neg c\end{smallmatrix}\right], 0.6\rangle\}$;

2. $i = 2$; $Rest = \left[\begin{smallmatrix}\neg c\\c, \neg c\\c, \neg c\end{smallmatrix}\right]$; $p_{Rest} = 0$;

3. $i = 3$; $Rest = \left[\begin{smallmatrix}\neg c\\c\\\neg c\end{smallmatrix}\right]$; $p_{Rest} = 0.2$, $E' = \left[\begin{smallmatrix}\neg c\\c\\\neg c\end{smallmatrix}\right]$; $p(E') = 0.2$;
   List = $\{\langle\left[\begin{smallmatrix}c\\c, \neg c\\c, \neg c\end{smallmatrix}\right], 0.6\rangle, \langle\left[\begin{smallmatrix}\neg c\\c\\\neg c\end{smallmatrix}\right], 0.2\rangle\}$. END

### 4.2 An algorithm for searching a conditional decision

The algorithm is sketched as Algorithm 3. Two steps of the algorithm deserve some comments: first (line 1), the sequences of CSP $\langle \Lambda \cup X, W \times D, C \cup E\rangle$, repeatedly solved by the algorithm, define *dynamic* constraint satisfaction problems [4] and their resolution may be improved by any techniques developed to solve such problems. Second (line 2), once a new $WC$ is built, it is subtracted from all the awaiting environments to avoid some redundant computations. This furthermore guarantees that the computation will stop when the current list Decisions defines a solution.

*Example (continued):*

1. $Env = \{\langle W, 1\rangle\}$; $p_{good} = 0$; $p_{bad} = 0$; Decisions = ∅

2. $\langle E, p_E\rangle = \langle W, 1\rangle$; $d_1 = (R, B)$; $WC = \left[\begin{smallmatrix}\neg c\\c, \neg c\\\neg c\end{smallmatrix}\right]$;
   $Env = \{\langle\left[\begin{smallmatrix}c\\c, \neg c\\c, \neg c\end{smallmatrix}\right], 0.6\rangle, \langle\left[\begin{smallmatrix}\neg c\\c\\\neg c\end{smallmatrix}\right], 0.2\rangle\}$;
   Decisions = $(\langle\left[\begin{smallmatrix}\neg c\\c, \neg c\\\neg c\end{smallmatrix}\right], (R, B)\rangle)$; $p_{good} = 0.2$.

3. $\langle E, p_E\rangle = \langle\left[\begin{smallmatrix}c\\c, \neg c\\c, \neg c\end{smallmatrix}\right], 0.6\rangle$; $d_2 = (R, T)$; $WC = \left[\begin{smallmatrix}c, \neg c\\c, \neg c\\\neg c\end{smallmatrix}\right]$;
   $Env = \{\langle\left[\begin{smallmatrix}c\\c\\c\end{smallmatrix}\right], 0.3\rangle, \langle\left[\begin{smallmatrix}c, \neg c\\c\\\neg c\end{smallmatrix}\right], 0.2\rangle\}$;
   Decisions = $(\langle\left[\begin{smallmatrix}c, \neg c\\c, \neg c\\\neg c\end{smallmatrix}\right], (R, T)\rangle, \langle\left[\begin{smallmatrix}\neg c\\c, \neg c\\\neg c\end{smallmatrix}\right], (R, B)\rangle)$;
   $p_{good} = 0.5$.

4. $\langle E, p_E\rangle = \langle\left[\begin{smallmatrix}c\\c\\c\end{smallmatrix}\right], 0.3\rangle$; $d_3 = (W, F)$; $WC = \left[\begin{smallmatrix}c, \neg c\\c, \neg c\\c, \neg c\end{smallmatrix}\right]$;
   $Env = \{\langle\left[\begin{smallmatrix}c\\c\end{smallmatrix}\right], 0.27\rangle, \langle\left[\begin{smallmatrix}\neg c\\c\end{smallmatrix}\right], 0.18\rangle\}$; $(\langle\left[\begin{smallmatrix}c, \neg c\\c, \neg c\\c, \neg c\end{smallmatrix}\right], (W, F)\rangle$ is added to Decisions; $p_{good} = 0.55$;

5. $\langle E, p_E\rangle = \langle\left[\begin{smallmatrix}c\\c\end{smallmatrix}\right], 0.27\rangle$; *inconsistent*; Bad = $\{\left[\begin{smallmatrix}c\\c\end{smallmatrix}\right]\}$;

```
Decisions := ∅           {decisions & env. covered}
Env := {⟨W, 1⟩}          {uncovered environment}
Bad := ∅                 {uncoverable envitonment}
p_good := 0              {lower bound of P_Cons(P)}
p_bad  := 0              {lower bound of 1 − P_Cons(P)}
repeat
   Pick a pair ⟨E, p_E⟩ from Env   {possible heuristics}
1  if ⟨Λ ∪ X, W × D, C ∪ E⟩ is inconsistent then
      % E contains only non-coverable worlds
      Bad := Bad ∪ E; p_bad := p_bad + p_E
   else
      (d, ω) := a solution of ⟨Λ ∪ X, W × D, C ∪ E⟩
      % d covers at least one possible world of E
      WC := C[d]                 {Worlds covered by D }
      Add ⟨WC, d⟩ to Decisions;
      Env' = ∅
      for G ∈ Env do
2        ⟨H, p⟩ := DEC(G, WC)
         Env' := Env' ∪ H
         p_good := p_good + p
      Env := Env'
until (Env = ∅) (or interruption by the user)
```

Algorithm 3: Computing an optimal conditional decision

$p_{bad} = 0.27$. Env = $\{\langle [\begin{smallmatrix}\neg c\\ c\end{smallmatrix}], 0.18\rangle\}$.

6. $\langle E, p_E\rangle = \langle[\begin{smallmatrix}\neg c\\ c\end{smallmatrix}], 0.18\rangle$; *inconsistent*;
   Bad = $\{[\begin{smallmatrix}c\\ c\end{smallmatrix}], [\begin{smallmatrix}\neg c\\ c\end{smallmatrix}]\}$; $p_{bad} = 0.45$; Env = ∅. END

$(\langle[\begin{smallmatrix}c,\neg c\\ c,\neg c\end{smallmatrix}], (W, F)\rangle, \langle[\begin{smallmatrix}c,\neg c\\ \neg c\end{smallmatrix}], (R, T)\rangle, \langle[\begin{smallmatrix}\neg c\\ \neg c\end{smallmatrix}], (R, B)\rangle)$ is the *final value of* Decisions; *from it we may built a conditional decision; moreover* $Bad$ *is in* Bad. $p_{good}$ *tends to* $P_{Cons}(P)$ *and reaches it eventually.*

### 4.3 Correctness of the algorithm

**Proposition 2** *At any point of the algorithm, the following properties hold:*

- $\forall \langle E, p_E\rangle$ *of* Env *we have* $p_E = Pr(E)$;

- $\forall E \in$ Bad *we have* $E \subseteq Bad(P)$;

- $\forall \langle E$ *and* $d\rangle \in$ Decisions, $\forall \omega \in E$, *d covers* $\omega$;

- $p_{good} \leq Pr(Good(P)) \leq 1 - p_{bad}$

**Proposition 3 (Correctness of the algorithm)** *If run quietly, the algorithm stops and :*

- *the final value of* Decisions *defines an optimal conditional decision;*

- *the final values of* $p_{good}$ *and* $p_{bad}$ *verify* $p_{good} + p_{bad} = 1$, $p_{good} = Pr(Good(C))$ *and* $p_{bad} = Pr(Bad(P))$.

Thus, the set of worlds covered by Decisions grows monotonically (and so does Bad), and the interval $[p_{good}, 1 - p_{bad}]$ shrinks monotonically from its initial value $[0, 1]$ to its final singleton value $[Pr(Good(C)), Pr(Good(C))]$. Therefore, it is possible to use the algorithm as an "anytime" algorithm: Decisions tends to a cover of $Good(P)$ and Bad tends to $Bad(P)$.

So far we have assumed that the applicability of a conditional decision is taken for granted (since, according to assumption (CK), the actual world will be revealed before the deadline for acting). We could consider more complex situations where further knowledge about the world can be learned by means of knowledge-gathering actions; another problem consists then in finding a relevant set of knowledge-gathering actions, sufficient to make the agent able to act properly. This is left for further research (see [5] for further considerations on this point).

## 5 Related work

*In the field of constraint satisfaction and automated reasoning*: Partial and valued CSP [6, 15] define general frameworks that extends the traditional CSP framework in a similar direction, but without any distinction between decision variables and parameters. Another related area of research is *dynamic constraint satisfaction* [4], where several techniques are proposed in order to solve a sequence of CSP that differs only in some constraints more efficiently than by naively solving each CSP one after the other. Such improvements could immediately be incorporated in the algorithm proposed. The computation of a compact representation of (partial) solutions could probably be performed using Finite Automata or Ordered Binary Decision Diagrams [2]. OBDD are used to solve problems issued from propositional logic, closely related to CSP. In fact, an OBDD could represent — using exponential memory and time in the worst case — all the partial solutions of a probabilistic CSP by shifting to propositional logic.

*In the field of decision analysis*: Deciding under uncertainty has been studied for long in decision theory and has been applied recently to planning. There is a number of computational approaches to decision analysis, which all distinguish in a way or another parameters (subject to probability distributions) and decision variables, the most common of them include influence diagrams [9], Markov decision processes and also valuation-based systems [16]. On the one hand, these computational frameworks tackle much more general decision problems since they consider *sequences of decisions* (and also utility values), while ours tackles only one-step decisions. On the other hand, our framework describes the constraints between dependent variables by means of an elaborate representation language equipped with elaborate computational tools (namely constraint satisfaction problems) while influence diagrams and other approaches do not focus on this representation issue and represent these relationships more explicitly. Both kind of approaches are however somewhat complementary, and one could think of extending our framework so as to integrate for instance influence diagrams (for the nice representation of the dependency structure of the decision problem) and constraint satisfaction (for the representation of the constraints be-tween dependent variables).

An interesting particular problem related to our framework is in Qi and Poole [13] who discuss a navigation problem in *U-graphs*, i.e., graphs where some of the edges are uncertain (the probability associated to an uncertain edge is the probability that the connection between two vertices is traversable); the static version of this problem may be encoded by a probabilistic CSP where a parameter is associated to each uncertain edge[4].

*In the field of knowledge representation*: Recently, Boutilier [1] has proposed a logical, qualitative basis for decision theory, distinguishing as we did between controllable and uncontrollable propositions; however, while we look for conditional decisions, Boutilier's very cautious strategy looks only for a pure decision (the best one maximizing the worst possible outcome). [12] also discusses controllability issues in an abduction-based framework for planning.

## 6 Concluding remarks

Our contribution was mainly in extending the CSP framework in order to deal with decision problems under uncertainty, by distinguishing between controllable variables and uncontrollable parameters and by representing the knowledge of the world by a probability distribution on the parameters. This framework can be considered as a first step to embed decision theory into constraint satisfaction; two other important steps in this direction will consists in considering *utilities* and *sequences of decisions*. Utility functions would enable us to represent *flexibility* and it should not be significantly harder to embed them in our framework: in the constraints of $C$, an extra field is added to each tuple, namely the utility of the corresponding variable assignment. Forbidden tuples have a utility of $-\infty$. Non-flexible constraints allow only two utility degrees for the tuples, namely 0 and $-\infty$. Extending our framework to sequences of decisions is significantly harder. The partition of the variables is not sufficient: not only decisions are influenced by parameters, but the value of some parameters may also be influenced by some earlier decisions; for this we think of reusing ideas from influence diagrams [9] or causal networks (*e.g* [3]) by structuring decision variables and parameters in a directed acyclic network (not to be confused with the undirected constraint graph of the CSP), a link from $\lambda_i$ to $x_j$ meaning that the allowed values of the decision variable $x_j$ depend on $\lambda_i$, and a link from $x_i$ to $\lambda_j$ meaning that the decision of assigning a value to $x_j$ has or may have some effects on $\lambda_j$.

An interesting potential application of probabilistic CSP (and its possible extensions) is *planning under uncertainty*. A preliminary version of probabilistic CSP has been applied to an agricultural planning problem [11]. Clearly, this problematic needs the notion of controllability, and would certainly gain a lot in being encoded in an extension of the CSP framework — since it could benefit from the numerous advances on the resolution of CSPs. The similarity between our conditional decisions and conditional plans is clear, at least for a single action. For handling sequences of actions, probabilistic constraint satisfaction has to be extended as evoked in the previous paragraph. Actions with probabilistic effects may be encoded by using extra parameters and probability distributions. Lastly, a lot of recent approaches to planning make use of decision theory; clearly, extending probabilistic CSP with utility values and sequences of decisions goes in this direction; our long-term goal is thus to provide a constraint satisfaction based framework to decision-theoretic planning.

---

[4]However the dynamic version of the problem, where the actual presence of an edge can be learned by the agent only when it is at one on its extremities cannot be easily modeled in our framework yet.